\documentclass[twoside,11pt]{article}

\usepackage[arxiv]{melba}

\usepackage{mwe} 

\usepackage{amsmath,amsfonts}
\usepackage{bm}

\newcommand{\modelname}{RKM}
\newcommand{\aff}{\bm{A}}
\newcommand{\moveaff}{\aff_m}
\newcommand{\fixaff}{\aff_f}

\newcommand{\moveimg}{\bm{I}_m}
\newcommand{\fiximg}{\bm{I}_f}
\newcommand{\alignimg}{\bm{I}_r}
\newcommand{\transformation}{\mathcal{T}}
\newcommand{\opttran}
{\transformation^{*}}

\newcommand{\nnparam}{w}
\newcommand{\nn}{f_\nnparam}
\DeclareMathOperator*{\argmax}{arg\,max}

\usepackage{pifont}
\usepackage{colortbl} 

\melbaid{YYYY:NNN}  
\doi{https://doi.org/10.59275/j.melba.2023-AAAA}
\melbaauthors{Moghadam, Wang, Taub, Prince, and Sabuncu}  
\volume{2}
\firstpageno{1}  
\melbayear{YYYY}  
\datesubmitted{m1/yyyy}  
\datepublished{m2/yyyy}  

\melbaspecialissue{Medical Imaging with Deep Learning (MIDL) 2020}
\melbaspecialissueeditors{Marleen de Bruijne, Tal Arbel, Ismail Ben Ayed, Hervé Lombaert}

\ShortHeadings{Resolution-agnostic Keypoint-based Registration}{Moghadam et al.}

\title{RealKeyMorph: Keypoints in Real-world Coordinates for Resolution-agnostic Image Registration}

\author{\firstname Mina \surname C. Moghadam* \email mch4003@med.cornell.edu \\  
	\addr Department of Radiology, Weill Cornell Medicine, New York City, NY, USA
	\AND
    \firstname Alan Q. \surname Wang* \email aw847@cornell.edu \\  
	\addr Department of Computer Science, Stanford University, Palo Alto, CA, USA
	\AND 
\firstname Omer \surname Taub* \email omt4002@med.cornell.edu \\  
	\addr Department of Radiology, Weill Cornell Medicine, New York City, NY, USA
	\AND
 \firstname Martin R. \surname Prince  \email map2008@med.cornell.edu \\  
	\addr Department of Radiology, Weill Cornell Medicine, New York City, NY, USA
	\AND
\name Mert R. Sabuncu \email msabuncu@cornell.edu \\
	\addr School of Electrical and Computer Engineering, Cornell Tech, New York City, NY, USA\\
    Department of Radiology, Weill Cornell Medicine, New York City, NY, USA
}

\begin{document}

\maketitle

\begin{abstract}
	%
Many real-world settings require registration of a pair of medical images that differ in spatial resolution, which may arise from differences in image acquisition parameters like pixel spacing, slice thickness, and field-of-view.
However, all previous machine learning-based registration techniques resample images onto a fixed resolution.
This is suboptimal because resampling can introduce artifacts due to interpolation.
To address this, we present RealKeyMorph ({\modelname}), a resolution-agnostic method for image registration.
\modelname~is an extension of KeyMorph~\citep{evan2022keymorph,wang2023keymorph}, a registration framework which works by training a network to learn corresponding keypoints for a given pair of images, after which a closed-form keypoint matching step is used to derive the transformation that aligns them.
To avoid resampling and enable operating on the raw data, \modelname~outputs keypoints \textit{in real-world coordinates} of the scanner.
To do this, we leverage the affine matrix produced by the scanner (e.g., MRI machine) that encodes the mapping from voxel coordinates to real world coordinates.
By transforming keypoints into real-world space and integrating this into the training process, \modelname~effectively enables the extracted keypoints to be resolution-agnostic.
In our experiments, we demonstrate the advantages of \modelname~on the registration task for orthogonal 2D stacks of abdominal MRIs, as well as 3D volumes with varying resolutions in brain datasets.
\end{abstract}


\section{Introduction}
Image registration is essential in medical image processing for aligning scans within or across subjects~\cite{maintz1998survey,sotiras2013deformable,fu2020deep,kurugol2017motion}. It plays a key role in longitudinal studies by tracking anatomical changes and disease progression over time and comparing disease severity between subjects. Additionally, it serves as a crucial preprocessing step in multimodal image analysis, aligning MR sequences or planes for segmentation and reconstruction.

Traditional, non-learning-based registration relies on iterative optimization of a similarity metric within a defined transformation space~\citep{oliveira2014medical,li2016image,sotiras2013deformable,chui2003new,hill2001medical}. Despite their effectiveness, these methods are computationally expensive, often requiring several minutes to register a single pair of images, and they tend to perform poorly when there is substantial initial misalignment between images. Among traditional approaches, keypoint-based methods have been widely explored as a means to improve robustness and interpretability~\citep{bay2008speeded,rosenfeld1971edge,Forstner1987,harris1988combined,Montesinos1998,wachinger2018keypoint}. By focusing on keypoints rather than dense voxel grids, keypoint-based methods reduce computational complexity, handle large misalignments robustly, and enhance interpretability. However, their effectiveness depends heavily on the quality of the detected keypoints, and they can struggle in cases with significant contrast or intensity variations~\citep{verdie2015tilde}.


In recent years, deep learning-based approaches have emerged as a promising alternative, leveraging large datasets to train neural networks for image registration~\citep{fu2020deep,haskins2020deep,cao2018deformable,eppenhof2018deformable,lee2019image,uzunova2017training,yang2017quicksilver,balakrishnan2019voxelmorph}. These techniques train neural networks to find the best transformation for each voxel or a set of representative keypoints in the moving images. The recently proposed KeyMorph technique~\citep{evan2022keymorph,wang2023keymorph} integrates deep learning into keypoint-based registration to detect an optimal keypoint set and compute transformations through a closed-form solution. This technique combines the robustness and interpretability of keypoint-based methods, particularly in handling large misalignments, with the fast inference times of deep learning. Additionally, KeyMorph supports various transformations based on user specifications, enhancing human controllability in the registration process.
A common characteristic across existing deep learning-based registration methods is their reliance on resampling paired images onto a standardized grid, often with isotropic resolution. While this simplifies computations, it reduces robustness when the input image pairs differ in dimensions or resolutions. This limitation is particularly pronounced when registering 2D MR stacks, specially orthogonal planes, with high slice thicknesses (greater than 2mm), where extensive interpolation can introduce inaccuracies. 

\textbf{Contribution.} In this work we present RealKeyMorph (RKM), as an extension of KeyMorph~\citep{evan2022keymorph,wang2023keymorph}, designed to compute keypoints in real-world coordinates. Our approach yields a resolution-agnostic deep learning-based registration model capable of handling images with varying input resolutions and dimensions. By learning keypoint correspondences in real-world coordinates, the model eliminates the need to resample images onto a common spatial grid, thereby removing an extra interpolation step, preserving the integrity of the original data, and improving overall accuracy. 
RKM enjoys the benefits of KeyMorph, such that the learned keypoints can be visualized to interpret what is driving the alignment.
RKM also offers users the flexibility to compute a dense set of registrations at test time directly from the learned keypoints. This feature allows users to control the level of non-linear deformation at inference time, tailoring it to the specific requirements of each application. 

The experimental analysis in this work is, in part, motivated by the application of image registration as a pre-processing step for high-resolution 3D volume reconstruction in abdominal MRI. Specifically, we utilize orthogonal 2D MRI stacks of abdominal images from 350 patients — comprising axial, coronal, and sagittal planes— which we aim to precisely align ~\citep{rousseau2006registration}. In this context, the in-plane resolution varies across scans, while the slice thickness introduces an out-of-plane resolution that is significantly coarser than the in-plane resolution. Additionally, patient movement during the acquisition of different planes adds further complexity to the registration process. Our resolution-agnostic RKM approach, tested on 140 pairs of axial and coronal 2D stacks from 50 patients, achieves a state-of-the-art median Dice score of 79\% and a Hausdorff Distance (HD) of 23 mm, underscoring the model’s ability to accurately register MRI planes with varying resolutions. 

To further explore the effectiveness and applicability of RKM, we performed a second analysis on the publicly available ADNI brain dataset, focusing on registration between image pairs with varying resolutions. Our goal was to include images from different MRI sequences to evaluate the model’s ability to generalize across diverse MRI acquisition protocols. The brain dataset included 3,480 pairs of 3D images with arbitrary MRI sequences, yielding registration performance of 89\%(4\%) Dice score and 48 mm (13 mm). These results demonstrate RKM’s capability to capture both affine and non-linear transformations between image pairs in the brain dataset. Our code is available at~\url{https://github.com/alanqrwang/keymorph}.

\section{Background}
Image registration is a rich literature that has been recently impacted by machine learning. Here we provide a very limited overview.

\textbf{Registration strategies.} Classical registration approaches involve pairwise iterative optimization techniques~\citep{hill2001medical,oliveira2014medical,li2016image}. While these methods perform well, they typically suffer from long runtimes, often requiring several minutes to register a pair of images. Another class of registration techniques, known as keypoint-based registration, involves extracting features—such as keypoints—from images and establishing correspondences between them. These features are often handcrafted or derived from image properties such as contours, intensity, color information, or segmented regions~\citep{bay2008speeded,rosenfeld1971edge,Forstner1987,harris1988combined,Montesinos1998,van2005boosting,matas2004robust,wachinger2018keypoint}. Algorithms then optimize similarity functions based on these features over the space of transformations~\citep{chui2003new,hill2001medical}. While keypoint-based methods are robust to initial misalignments and offer improved interpretability, their effectiveness depends heavily on the quality of the extracted keypoints and can degrade under significant contrast or color variations~\citep{verdie2015tilde}.

More recently, deep learning-based registration strategies have emerged, where neural networks are trained to perform the registration task. These strategies use convolutional neural networks (CNNs) or transformer-based architectures to output either transformation parameters or a deformation field that aligns image pairs. The training process can be supervised—using existing ground-truth data~\citep{cao2018deformable,dosovitskiy2015flownet,eppenhof2018deformable,lee2019image,uzunova2017training,yang2017quicksilver}—or unsupervised, by optimizing loss functions similar to those used in classical methods~\citep{balakrishnan2019voxelmorph,dalca2019unsupervised,de2019deep,fan2018adversarial,krebs2019learning,qin2019unsupervised,wu2015scalable,hoopes2021hypermorph,wang2024recursive}. Optimization techniques such as gradient descent, or more advanced methods like Adam, are employed to minimize the chosen similarity metric and iteratively refine the alignment.

Furthermore, recent learning-based methods have extended this approach by learning useful features or keypoints, followed by establishing correspondences between image pairs—typically by identifying the most similar learned features across the pair~\citep{verdie2015tilde,yi2016lift,yi2018learning,detone2018superpoint,liu2021same,barroso2019key,lenc2016learning,ono2018lf}. For example, the recently proposed KeyMorph technique~\citep{evan2022keymorph,wang2023keymorph} integrates keypoint-based registration with deep learning by automatically detecting corresponding keypoints using neural networks, which generate keypoints directly from the image in a closed-form manner. These keypoints enable the computation of optimal transformations analytically, combining the robustness and interpretability of keypoint-based methods with the fast inference times of deep learning.
 
\textbf{Loss functions and similarity metrics.} Registration methods employ various similarity metrics and loss functions to achieve precise image alignment. Intensity-based metrics like mean-squared error (MSE) and normalized cross-correlation are commonly used for images of the same modality, measuring the pixel-wise similarity between images~\citep{avants2009advanced,avants2008symmetric,Hermosillo2002Variational}. For multimodal image registration, statistical measures such as mutual information (MI) and the modality independent neighborhood descriptor (MIND) have been used~\citep{Heinrich2012MIND,Hermosillo2002Variational,Hoffmann2021Synthmorph,Mattes2003PET-CT,viola1997alignment}. In addition to the moving and fixed image, a supplementary image, e.g., in the form of segmentation masks can help to drive the alignment, where, for example, the overlap of the segmentation masks can be an additional metric to optimize the registration process~\citep{balakrishnan2019voxelmorph,Hoffmann2021Synthmorph,Zhang2021Two,Song2022Cross}. However, the process of generating these segmentation masks is usually time-consuming and expensive.
 
\textbf{Spatial Transformations.} Deformation models play a crucial role in registration methods. These models define how an image can be warped or transformed to match another. Common deformation models include linear transformations i.e. rigid, affine, and non-linear transformations, each offering varying degrees of flexibility. Rigid transformations involve translation and rotation, where  Ecludian distances are preserved. Affine transformations which are the simplest form of non-rigid transformation include scaling and shearing as well.  These linear deformations, however, cannot capture most organ deformations and local movements. Non-linear deformable transformations are more suitable for these cases, often with diffeomorphic constraints to preserve the topology of the registered structures~\citep{dalca2018MICCAI,dalca2019unsupervised}. Transformation constraints and regularization strategies are essential to maintaining biologically plausible deformations and preventing overfitting.  
 
Transformation models are chosen based on the underlying problem and the type of organs they aim to align. 
In some applications, e.g., alignment of a moving fetal brain, rigid transformations might be sufficient~\citep{xu2022svort,rousseau2006registration}.
When dealing with brains of diffrent individuals, however, a common approach is to use non-rigid transformations with minimal deformation~\citep{balakrishnan2019voxelmorph,Hoffmann2021Synthmorph,wang2023keymorph,kang2022dual,ni2020robust}.  More flexible non-rigid transformations are often required for other domains, such as the heart, thoracic, and abdominal regions~\citep{ebner2017point,bhatta2020evaluation,Xu2021f3rnet,carrillo2000semiautomatic,xu2016evaluation,lei2024diffeomorphic,lv2018respiratory,kurugol2017motion,Tanner2022Generative,bhatta2020evaluation}. 
While the KeyMorph model focused on the registration of brains between pairs of subjects, here we aim to explore how a KeyMorph-based framework can perform in more complex situations, such as abdominal registration tasks. In Section 4, we will introduce the abdominal dataset we use for this purpose. 

\section{Methods}

Our method, RealKeyMorph ({\modelname}), is based on the KeyMorph (KM) framework, and we refer the reader to prior papers for more details~\citep{evan2022keymorph,wang2023keymorph}. 
For each image $\bm{I}(\cdot)$\footnote{Without loss of generality, we assume the image is a 3D volume or a stack of 2D slices} (e.g. MRI volume or 2D stack), we have access to a corresponding affine matrix $\aff$ (often available in the header, e.g., as a DICOM tag), which encodes the mapping from the voxel index $\bm{x}$ to real-world coordinates of the imaging device $\bm{x}_{real}$ (typically in millimeters): $\bm{x}_{real} = \aff \bm{x}$, where we have used homogeneous coordinate notation for mathematical convenience.  
Let $(\moveimg, \fiximg)$ be a moving (source) and fixed (target) image pair, possibly of different contrasts. As we intend to register orthogonal views of the same subject, our objective is to correct for organ and subject motion between these views. In this paper, we therefore consider non-rigid transformations, which will be modeled with thin-plate splines (TPS), as in~\citep{wang2023keymorph}.

Our goal is then to find the optimal spatial transformation $\opttran$ such that the registered image~$\alignimg = \moveimg \circ \opttran$ aligns with some fixed image~$\fiximg$, where~$\circ$ denotes the spatial transformation of an image. 
In other words, for images with the same intensity contrast, our objective can be considered as $\moveimg ( \aff_m^{-1} \opttran (\aff_f \bm{x})) \approx \fiximg (\bm{x})$, where $\aff_f$ and $\aff_m$ are the real-world transformation matrices for the fixed and moving images, respectively.
By working in real-world coordinates, our model can be resolution-agnostic, eliminating the need for interpolation and making the registration effective for image pairs with different intra-slice resolutions and slice thicknesses, as in the case of orthogonal views. 


\begin{figure*}[ht]
\centering
\includegraphics[width=\linewidth, height=0.29\textheight]{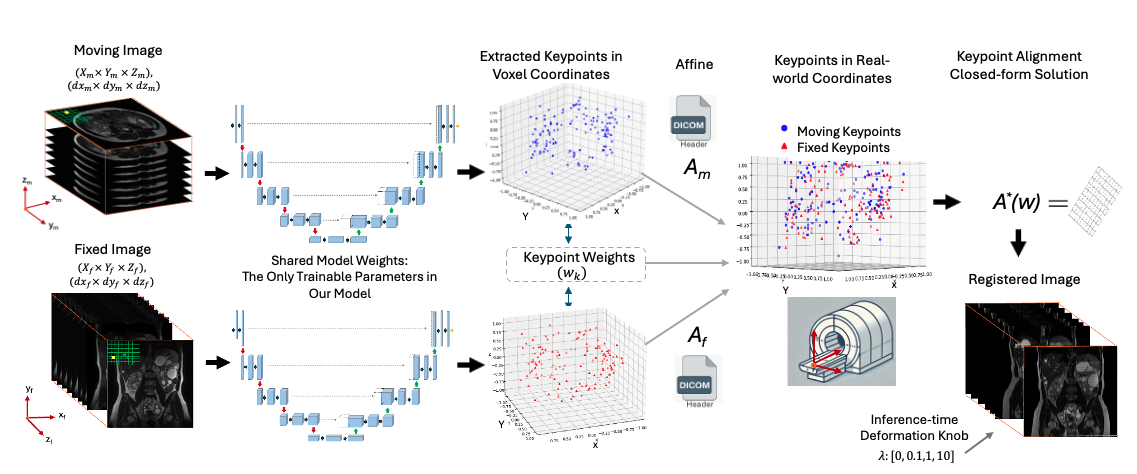}
\caption{Graphical overview of RKM.}
\label{fig:arch}
\end{figure*}

Fig.~\ref{fig:arch} depicts a graphical overview of our model. {\modelname} works by detecting $N \ge 4$ keypoints \textit{in real-world coordinates} $\{\bm{k}_i\}_{i=1}^{N}$ from any given image.
The keypoints are detected in a two-step procedure; first, a neural network $\nn$, with learnable network parameters $w$, outputs $N$ keypoints in voxel coordinates.
Second, these keypoint voxel-coordinates are transformed into real-world coordinates by matrix multiplication via the images' corresponding affine matrix.
Each keypoint is also associated with a confidence weight $c_i > 0$, computed as an output of the neural network. 
This captures how confident the model is in localizing a specific keypoint in the image and we found it to be critical in handling missing keypoints due to unusual pathologies or cropped fields of view. Since $\nn$ detects keypoints for any image, the keypoints for an arbitrary image pair will be in correspondence by construction.
We can then use these to derive the optimal transformation that minimizes the distance between the corresponding keypoints in real-world coordinates:
\begin{equation}
    \opttran(w) = \arg \min_{\transformation} \sum_{i=1}^N c_i^m(w) c_i^f(w) \|\bm{k}_{i}^{f}(w) - \bm{A}\bm{k}_{i}^{m}(w)\|^2
    \label{eq:wlss}
\end{equation}
where the super-scripts $m$ and $f$ indicate the moving and fixed image, respectively; and $(w)$ explicitly denotes the dependency between the computed keypoints and confidence weights, and the neural network parameters.
In Equation~\ref{eq:wlss}, $\| \cdot \|$ denotes Euclidean distance in real-world coordinates (often measured in millimeters). 
Note that the distance for each keypoint pair is weighted by the product of the confidences of these keypoints.

If we restrict our analysis to affine transformations, there is a closed-form solution for the weighted least squares problem of Equation~\ref{eq:wlss}. Let's denote this optimal affine transformation matrix as $\opttran_A(w)$:
\begin{equation}
    \opttran_A(w) = \bm{K}^f(w) \bm{C}(w) (\bm{K}^m(w))^T (\bm{K}^m(w) \bm{C}(w) (\bm{K}^m(w))^T)^{-1},
    \label{eq:opt}
\end{equation}
where $\bm{C}(w) = \textrm{diag}(c_1^m(w) c_1^f(w), \ldots, c_N^m(w) c_N^f(w))$ is a diagonal matrix of confidence weight products; and $\bm{K}^f(w) = [\bm{k}_{1}^{f}(w), \ldots,  \bm{k}_{N}^{f}(w)]$ and $\bm{K}^m(w) = [\bm{k}_{1}^{m(w)}, \ldots,  \bm{k}_{N}^{m}(w)]$ are the matrices of stacked keypoints from the fixed and moving images respectively.

A special case of the above formulation (i.e., an ablation of the proposed model) ignores the confidence weights, in which case the neural network only computes keypoint coordinates and $\bm{C}$ is an identity matrix.

As discussed in the original KM paper~\citep{wang2023keymorph}, we can relax the affine transformation assumption and consider non-linear transformations via the TPS model.
In TPS, there is a user-defined hyper-parameter $\lambda \ge 0$ that controls the degree of non-rigidity.
For large $\lambda$ values, TPS approaches affine transformations, whereas for $\lambda = 0$, corresponding keypoints are matched exactly, while interpolating the rest of the deformation to minimize so-called bending energy. 
For a fixed $\lambda$ parameter, similar to the affine case, there is a closed-form solution for the optimal non-rigid transformation.
We refer the reader to the original TPS and KeyMorph papers~\citep{bookstein1989principal,wang2023keymorph} for the mathematical details.
In the present paper, we train the model with both affine and TPS transformations.
At test time, the model can produce registrations for rigid, affine or non-rigid (TPS). 


\subsection{Keypoint Detection Network}
{\modelname} can leverage any deep learning-based keypoint detector.
In this work, we are interested in preserving translation equivariance; to this end, we leverage a center-of-mass (CoM) layer~\citep{ma2020volumetric, sofka2017fully} as the final layer, which computes the center-of-mass for each of the $N$ activation maps over the normalized space $[-1, 1]$, after which these centers-of-mass are converted to keypoints in voxel coordinates by scaling them according to the dimensions of the volume along each dimension. 
The reason keypoints are extracted in a normalized space is for optimization purposes.
This specialized layer is (approximately) translationally-equivariant and enables precise localization. 
Since the CoM layer expects positive values at every grid location, we insert a ReLU activation before the CoM layer.
The neural network also outputs confidence weights for each detected keypoint. 
In our implementation, these weights are computed as the spatial pooling of the corresponding non-negative activation map. 
Therefore, the model learns to produce lower/higher activation values for keypoints it is less/more confident about. Our model architecture is a simple 5-layer UNet with skip connections. Each convolutional layer is followed by instance normalization, ReLU activations, and strided convolutions for 2x downsampling. The input layer has one channel for the grayscale MRI of the moving or fixed image, while the output layer contains 4×N values for keypoint (x, y, z) locations and confidence scores.

\subsection{Training}
Training {\modelname} optimizes the learnable parameters of the CNN $\nn$ for pairwise registration:
\begin{equation}
\argmax_\nnparam \ \mathbb{E}_{(\moveimg, \fiximg)} \  \mathcal{L}_{sim}\left(\moveimg \circ \moveaff^{-1} \circ \opttran (w) \circ \fixaff, \fiximg\right)  
\label{eq:objective}
\end{equation}
where $\mathcal{L}_{sim}(\cdot, \cdot)$ measures the similarity between its two image inputs, $\circ$ denotes the concatenation of functions, and $\opttran (w)$ is defined in Equation~\ref{eq:opt}.
Note that we compute the similarity objective in the voxel coordinates of the fixed image, which does not have to be resampled. 
Thus, only the moving image pixel values need to interpolated on the fixed image voxel grid, via the concatenation of spatial transformations that map through the keypoint correspondence in the real-world coordinate frame.
Finally, we are free to define the similarity objective any way that is appropriate. 

In our experiments on the abdominal dataset, where we register between orthogonal views, we focus on images with the same contrast and assume the availability of segmentation labels to help guide the registration.
Therefore, our similarity objective is based on intensity-based metrics such as the negative of the sum of squared pixel intensity differences, or structural similarity index measures (SSIM)~\cite{wang2004image}, combined with overlap-based metrics such as the Dice score of segmentation labels. For the brain dataset, we extend this setup to register 3D volumes with varying MR sequences. As a result, we limit the loss function to consider only intensity-independent metrics, such as the Dice similarity metric, due to the differences in image contrast.
Note that, as in KM, we do not rely on any ground truth keypoints as supervision.
Instead, the model learns to detect keypoints and assign appropriate confidence weights in order to maximize the registration accuracy.


\subsection{Pairwise Registration: Inference}
Once we have a model trained on pairs of images, we can use it to perform registration on a new pair of images by running it in inference mode.
Note that inference is not restricted to a specific transformation type.
As discussed in the original KeyMorph paper~\citep{wang2023keymorph},
the pre-trained model computes keypoints and their corresponding confidence values, which can then be used to drive either rigid or non-rigid transformations by adjusting a user-defined hyperparameter, $\lambda$, that controls the amount of deformation.



\section{Experimental Setup}
\subsection{Datasets}
 In this study, we trained our models on two different datasets — brain and abdominal imaging — to assess their effectiveness in different organ settings. We included patients with different imaging dimensions, in-plane resolutions, and slice thicknesses to demonstrate the models' effectiveness in resolution-agnostic registration.

\textbf{Abdominal MRI Dataset:} We utilized abdominal MRI data from 206 patients (52 ± 12 years), collected at Weill Cornell Medicine and its affiliated Rogosin Institute~\citep{zhu2024primer}. Among these patients 180 have mild to moderate Autosomal Dominant Polycystic Kidney Disease (ADPKD), the most common inherited kidney disorder which frequently impacts other abdominal organs, including the liver and spleen, which affects imaging contrast and alters the anatomy of these organs. Imaging was performed on 1.5T or 3T MRI scanners from GE Healthcare and Siemens Healthineers. Due to the limitations of breath-holding in abdominal imaging, acquiring full 3D volumes is often impractical. As a result, the images are typically acquired as 2D MRI stacks, usually with a higher slice thickness than the in-plane resolution.  Most patients underwent multiple MR sequences, including T2-weighted, and Steady-State Free Precession (SSFP), captured in both axial and coronal planes. From 860 image pairs in our dataset (430 axial–coronal and 430 coronal–axial), 688 pairs (170 patients) were used for training, and 172 pairs (36 patients) were reserved for testing. The in-plane image resolution ranged from 0.59 to 1.87 mm (mean (std): 1.14 (0.29) mm) in axial stacks and 0.58 to 4.40 mm (mean (std): 1.62 (0.61) mm) in coronal stacks, with slice thickness varying from 0.82 mm to 10 mm (mean (std): 5.71 (1.92) mm). We also used the corresponding segmentation masks for the left kidney, right kidney, spleen, and liver generated by our in-house segmentation tool~\citep{sharbatdaran2022deep} to assess the training of the registration tasks, as well as for evaluation purposes. Our models use DICOM MRI sequences, and NIFTI segmentation masks as inputs.
Z-score normalization~\citep{shinohara2014normalization} standardizes DICOM intensities to ensure consistency across images.

\textbf{ADNI Brain Dataset:} We additionally evaluated our method on the ADNI brain dataset~\citep{petersen2010alzheimer}, which provides multi-modal 3D MRI scans for each subject, including T1-weighted, T2-weighted, and FLAIR sequences. Our cohort consisted of 665 subjects, with voxel resolutions ranging from 0.98 mm to 1.03 mm (mean ± std: 1.00 ± 0.01 mm) in all spatial dimensions. Image dimensions varied between 154×230×175 and 256×256×256 across subjects. For anatomical segmentation, we employed a pre-trained SynthSeg model~\citep{billot2023synthseg} to automatically delineate 32 brain regions. These segmentation masks enabled intensity-independent evaluation using overlap-based metrics such as the Dice coefficient, facilitating multi-modal registration without relying on intensity-based similarity measures. We generated a total of 3,479 intra-subject registration pairs, including both intra- and inter-modality combinations. From these, 2,742 pairs were used for training and 737 for validation.

\subsection{Model and Training Details}

We trained several variants of RKM to investigate the effects of different model configurations and the similarity objectives in Eq. \ref{eq:objective}. 
We evaluated image similarity using the mean squared error (MSE), structural similarity index measurements (SSIM), and their combinations with the Dice score and HD. To further improve performance, we analyzed the impact of model capacity by varying the number of layers (4, 5, or 6), ultimately identifying the 5-layer network as the most effective configuration. The models were trained to detect 32, 64, 128, or 256 keypoints. To mitigate the effect of having different fields of view between orthogonal MR stacks, the key-point detector model also assigns a confidence to each keypoint, accounting for its individual contribution to the registration task and thus allowing the possibility for missing or unreliable keypoints. Training used a batch size of one image pair, and a learning rate of $10^{-3}$, and the Adam optimizer for 4,000 iterations, where the loss plateaued. For TPS transformations, log-uniform sampling was applied to select lambda values ranging from 0.001 to 100, allowing for varying degrees of nonlinear deformation.


\textbf{Baseline.} We are not aware of any existing deep learning-based registration approaches which are agnostic to resolution and can work in the native coordinate space of the images, while accommodating differing resolutions.
Thus, as baselines we select two state-of-the-art registration approaches, VoxelMorph~\citep{balakrishnan2019voxelmorph} and KeyMorph~\citep{wang2023keymorph}, both of which depend on resampling all images to 256x256x256 1 $mm^3$ isotropic resolution before passing them into the registration models.

\subsection{Metrics}
We evaluate registration performance using SSIM, Soft Dice Score, and HD. For each subject, we compute the average Soft Dice Score and HD between the segmentation masks. To ensure a fair comparison between baseline methods and our proposed model, we resample the registered images from a $256^3$ voxel resolution back to the original fixed image resolution before evaluation.

\section{Results}

In this section, we present the experimental results demonstrating RKM's performance and its ability to handle medical images with varying resolutions. 

\subsection{RKM for Orthogonal 2D MRI Stacks}

We evaluate the performance of RKM on orthogonal 2D MR stacks within each MR sequence (e.g. T1, T2, SSFP) from our abdominal MRI dataset. The model was trained and tested using both affine and log-uniform TPS transformations across varying numbers of keypoints (32, 64, 128, and 256). For each configuration, we trained models with different loss functions that integrate both intensity-based metrics and Dice scores derived from segmentation masks. Given the limited availability of segmentation labels in our abdominal dataset, i.e. the liver, kidneys, and spleen, we combined Dice loss with intensity-based losses to enhance anatomical guidance during training. 

\begin{figure*}[t]
\centering
\includegraphics[width=1\linewidth, height=0.8\linewidth]{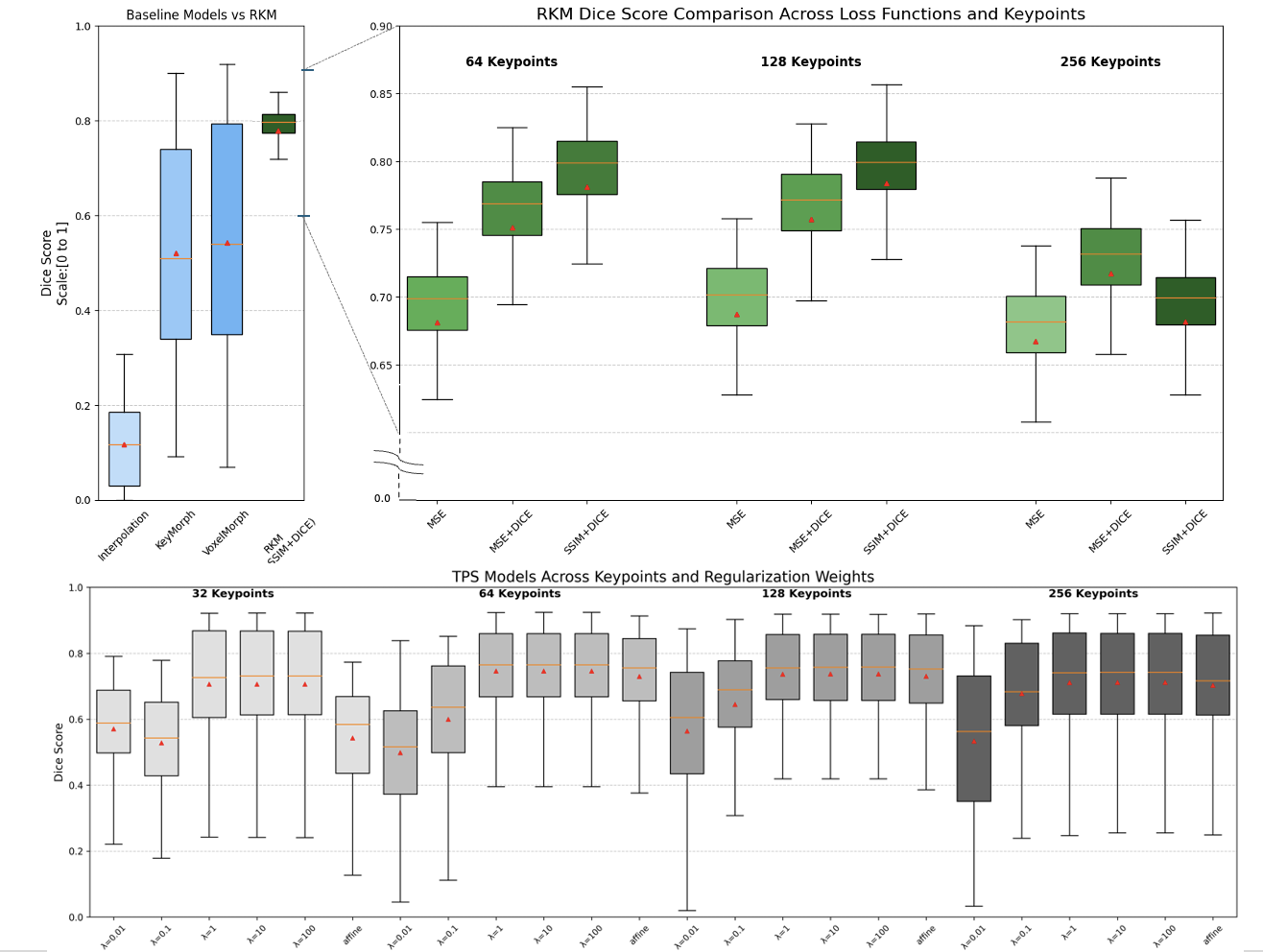}
\caption{Performance of the RealKeyMorph (RKM) model on the Abdominal dataset across different loss functions, number of keypoints, and transformation models.
Top Left: Comparison of baseline models (Interpolation, KeyMorph, VoxelMorph) with the best-performing RKM affine model using 64 keypoints and SSIM+DICE loss.
Top Right: Dice score distributions for RKM affine models using 64, 128, and 256 keypoints with three loss functions: MSE, MSE+DICE, and SSIM+DICE.
Bottom Panel: TPS-based RKM models evaluated across four keypoint configurations (32, 64, 128, 256) and five regularization weights ($\lambda = 0.01$ to $100$), plus affine-only variants.
Each box plot shows the interquartile range (IQR) as the width of the box, which reflects the spread of the middle 50\% of the data. The horizontal orange line represents the median, and the red triangle indicates the mean Dice score.
The best performance is observed for the affine RKM model with 64 keypoints and SSIM+DICE loss, yielding a median Dice score of 79\% and an average Dice score of 77\%.}
\label{fig:Dice Abdominal}
\end{figure*}

Fig.~\ref{fig:Dice Abdominal} (upper right panel) illustrates the RKM model's performance across different configurations. The results indicate that the best performance is achieved when the model optimizes SSIM+Dice, with a median (IQR) of 79\% (7\%) for the affine transformation.  Our observations suggest that 64 or 128 keypoints are optimal for our application. While lower numbers (e.g., 32) showed similar performance, increasing the number of keypoints beyond 128 leads to redundancy without improving registration performance as the keypoints tended to overlap with existing ones or cluster near the center of the image. Fig.~\ref{fig:Dice Abdominal} also shows the results for training TPS models (lower panel). We utilized the SSIM+Dice loss function to train a range of TPS registration models using log-uniform sampling across lambda values from 0.001 to 100, allowing for a range of non-linear deformation. It is shown that, the best performance was observed with a median (IQR) Dice score of 75\% (15\%) for 64 keypoints and a lambda value of 10.

\begin{table}[htbp] 
    \centering
    \caption{Similarity metrics for the RKM and two baseline models on abdominal dataset, Median (IQR).}
    \label{tab:baseline}
    \begin{tabular}{lllll}
        \textbf{Model} & \textbf{Keypoints} & \textbf{Dice} & \textbf{HD (mm)} & \textbf{SSIM} \\
        \hline
        VoxelMorph Baseline & NA & 54\% (44\%) & 49 (12) & 55\% (10\%) \\
        KeyMorph Baseline & 512 & 51\% (40\%) & 99 (40) & 44\% (14\%) \\
        RealKeyMorph (Affine) & 64 & 79\% (7\%) & 23 (10) & 60\% (10\%) \\
        RealKeyMorph (TPS) & 64 & 75\% (15\%) & 24 (12) & 64\% (12\%) \\

    \end{tabular}
\end{table}

Fig.~\ref{fig:Dice Abdominal} (upper left panel) also shows the initial Dice score before registration in voxel coordinates, as well as the performance of KM and VoxelMorph on standardized grid coordinates, compared with RKM, which operates on original, non-resampled images. Table~\ref{tab:baseline} summarizes that RKM outperforms baseline models, which is attributed to its ability to support resolution-agnostic registration, avoiding resampling and interpolation artifacts.

\begin{figure*}[t]
\centering
\includegraphics[width=0.85\linewidth, height=0.4\linewidth]{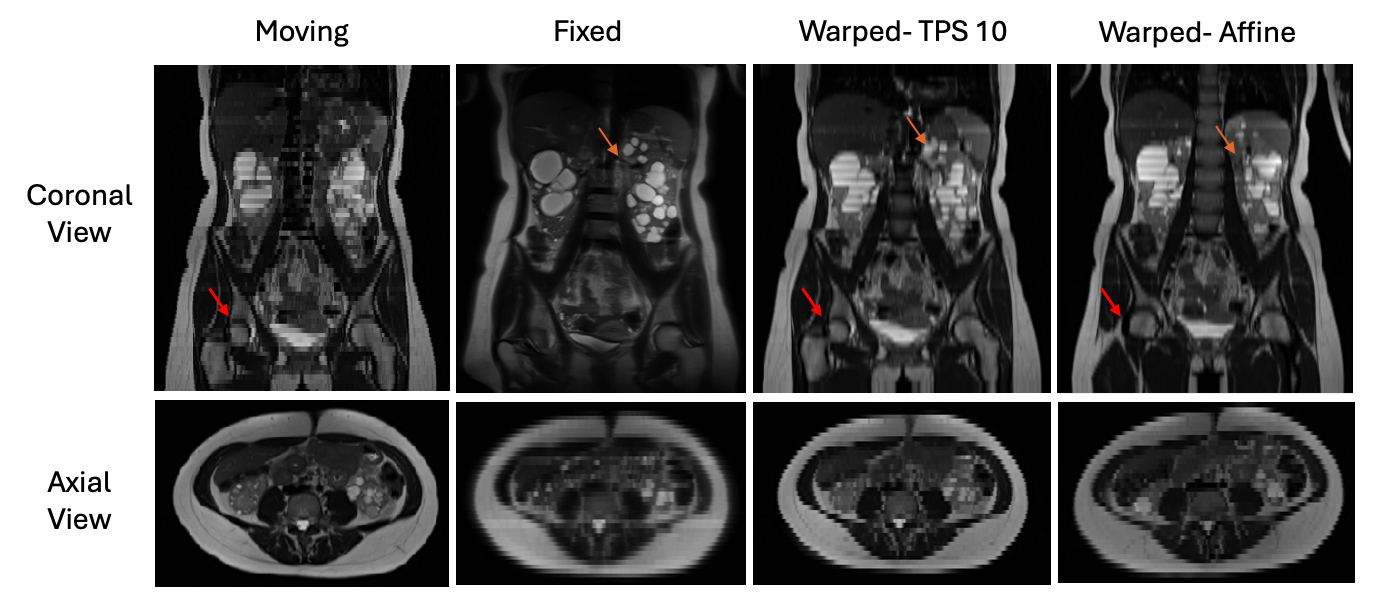}
\caption{Representative registration results obtained with proposed RKM, from various samples in the abdominal dataset, shown on orthogonal view pairs. The moving image (axial plane) has dimensions of 320, 240, 72, a resolution of 1.18 mm, 1.18 mm, and a slice thickness of 6 mm. The fixed image (coronal plane) has dimensions of 320, 42, 320, a resolution of 1.4 mm, 1.4 mm, and a slice thickness of 5 mm. Registered images are obtained using TPS ($\lambda=10$) and affine transformations. As indicated by the red arrows, the TPS-10 experiment appears to better capture anatomical details compared to the Affine model.}
\label{fig:Samples Abdominal}
\end{figure*}

Fig.~\ref{fig:Samples Abdominal} illustrates the registration results on a test set, aligning moving images in the axial plane with dimensions (320, 240, 72) to fixed images in the coronal plane with dimensions (320, 42, 320). The voxel spacing for the moving and fixed images is (1.18 mm, 1.18 mm, 6 mm) and (1.4 mm, 5 mm, 1.4 mm), respectively. The results are shown for the affine and TPS-10 experiments, as TPS-10 achieved the highest Dice score among the TPS experiments. The figure shows that, while the affine model reaches a higher Dice score (83\%) compared to TPS-10 (79\%) for this example, the TPS-10 experiment appears to better capture anatomical details. 

\begin{figure*}[t]
\centering
\includegraphics[width=0.65\linewidth, height=0.4\linewidth]{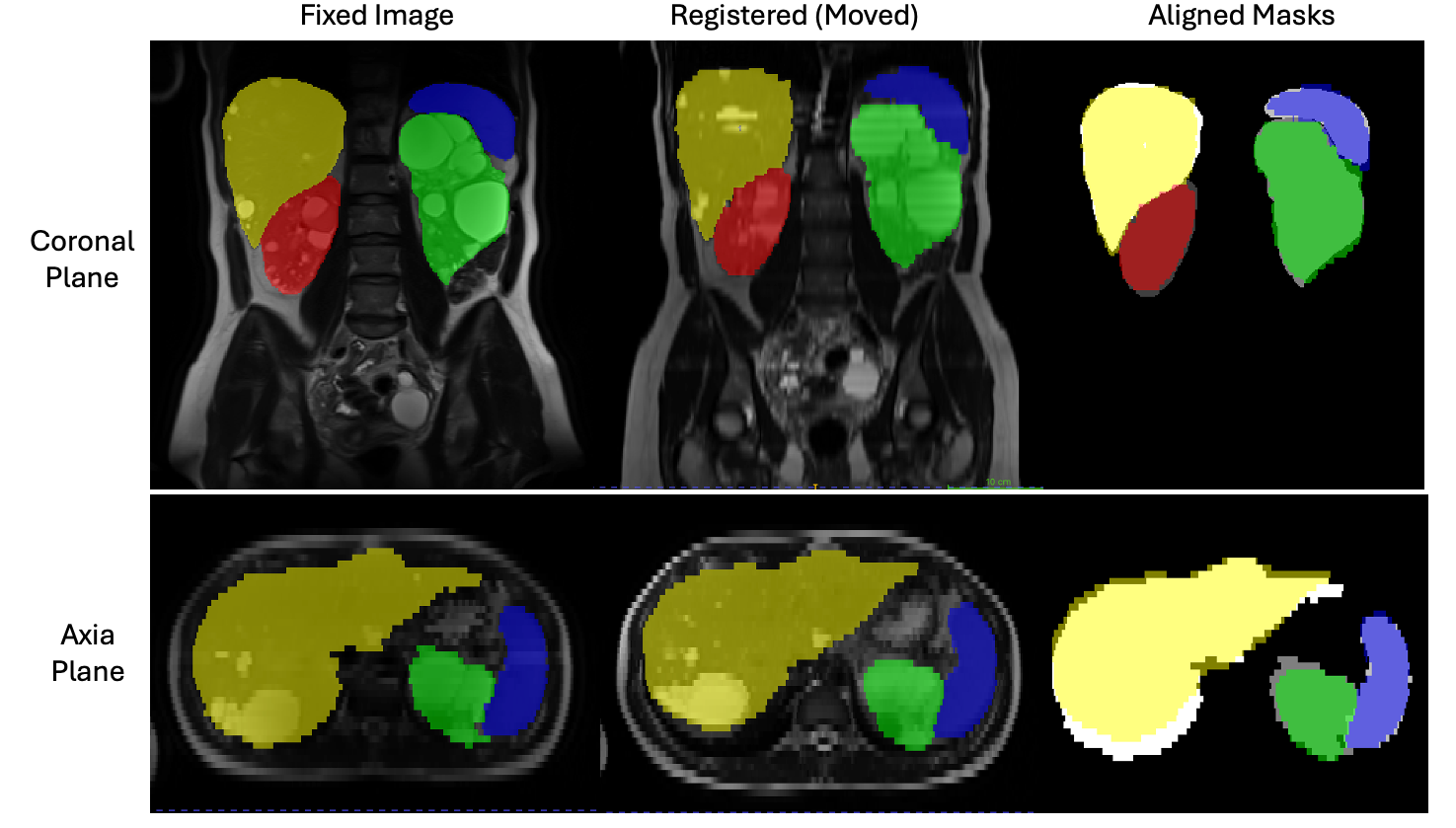}
\caption{Fixed and moved image as well as their mask alignments for representative slices in axial and coronal views from a sample patient in the test set. Red: Right Kidney, Green: Left Kidney, Yellow: Liver, and Blue: Pancreas.}
\label{fig:Masks}
\end{figure*}

Fig. \ref{fig:Masks} illustrates a registration example of
a moving image in axial plane to the fixed image in coronal plane, along with their segmentation masks. The right column shows the overlap between the segmentation masks after registration. Our results shows that, on average, across patients, RKM achieves the following median (IQR) Dice scores of 82\% (6\%),—83\% (6\%), 70\% (14\%), and 80\% (6\%), for Right Kidney, Left Kidney, Spleen, and Liver, respectively.

\begin{figure*}[t]
\centering
\includegraphics[width=0.61\linewidth, height=0.6\linewidth]{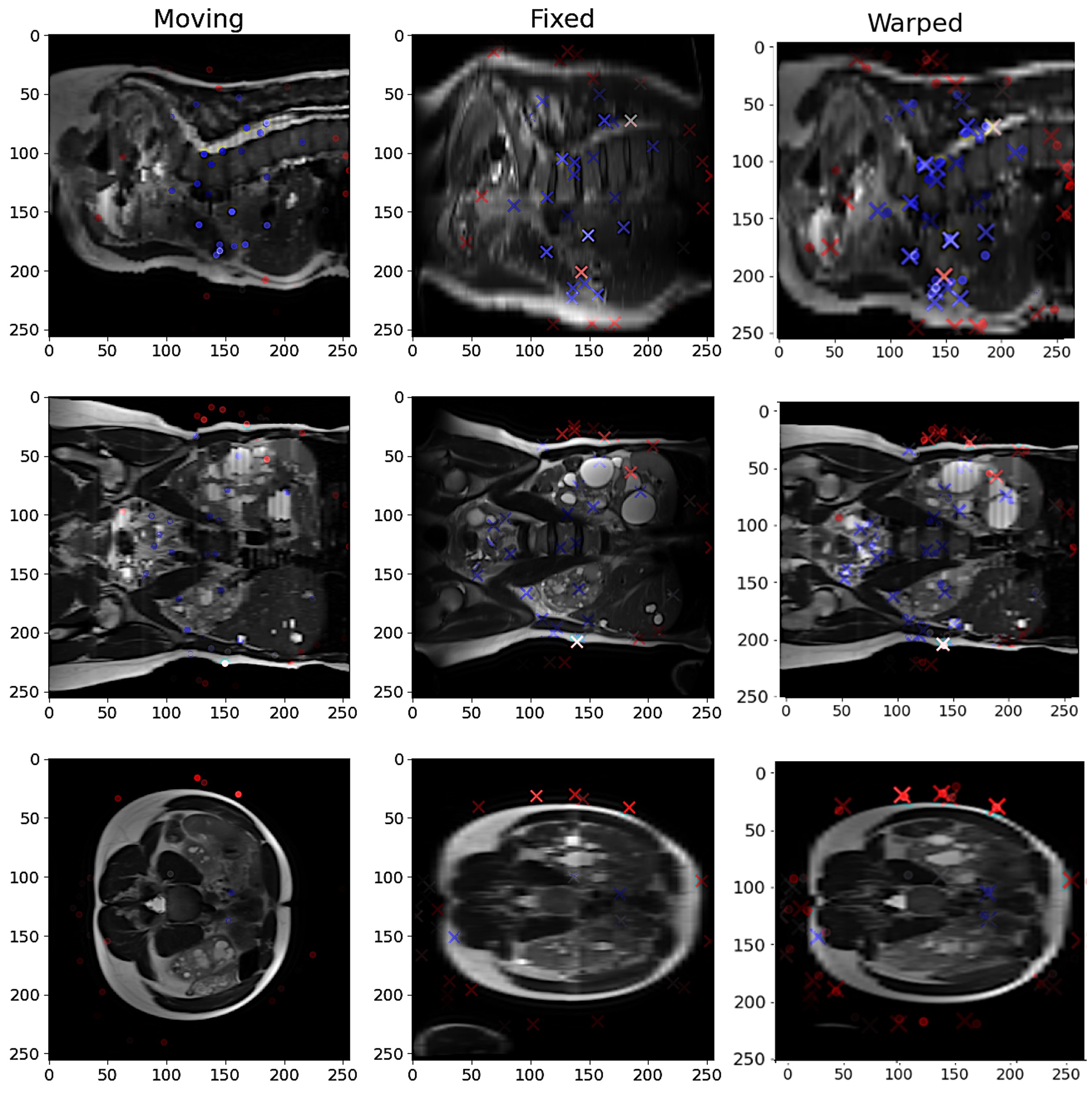}
\caption{Representative slices across three views for the moving, fixed, and warped images, along with the keypoints within \(\pm 1\) slice. The colors indicates keypoint distances from the visualized slice, while their transparency represents the keypoint weights.}
\label{fig:KeyPoints}
\end{figure*}

Fig.~\ref{fig:KeyPoints} shows a representative example of keypoint extractions on orthogonal T2-weighted MR sequence sample, where the moving and fixed images are in axial and coronal planes respectively. We observe that the model learns keypoints that consistently correspond to matching locations in the abdomen, regardless of the resolution of the volume across the three views.

\subsection{RKM for Multi-Modal Registration}

In this section, we evaluate the generalizability of RKM's resolution-agnostic registration using the widely available ADNI brain dataset. Additionally, we analyzed the multi-modal registration capabilities of RKM by registering images from different sequences within each subject in the ADNI dataset. Using segmentation masks covering 32 distinct brain regions ~\citep{billot2023synthseg}, we employed Dice loss as the primary training objective. Note that we restrict our analysis to affine registrations, given that this is sufficient for intra-subject brain alignment.

\begin{figure*}[t]
\centering
\includegraphics[width=0.95\linewidth, height=0.45\linewidth]{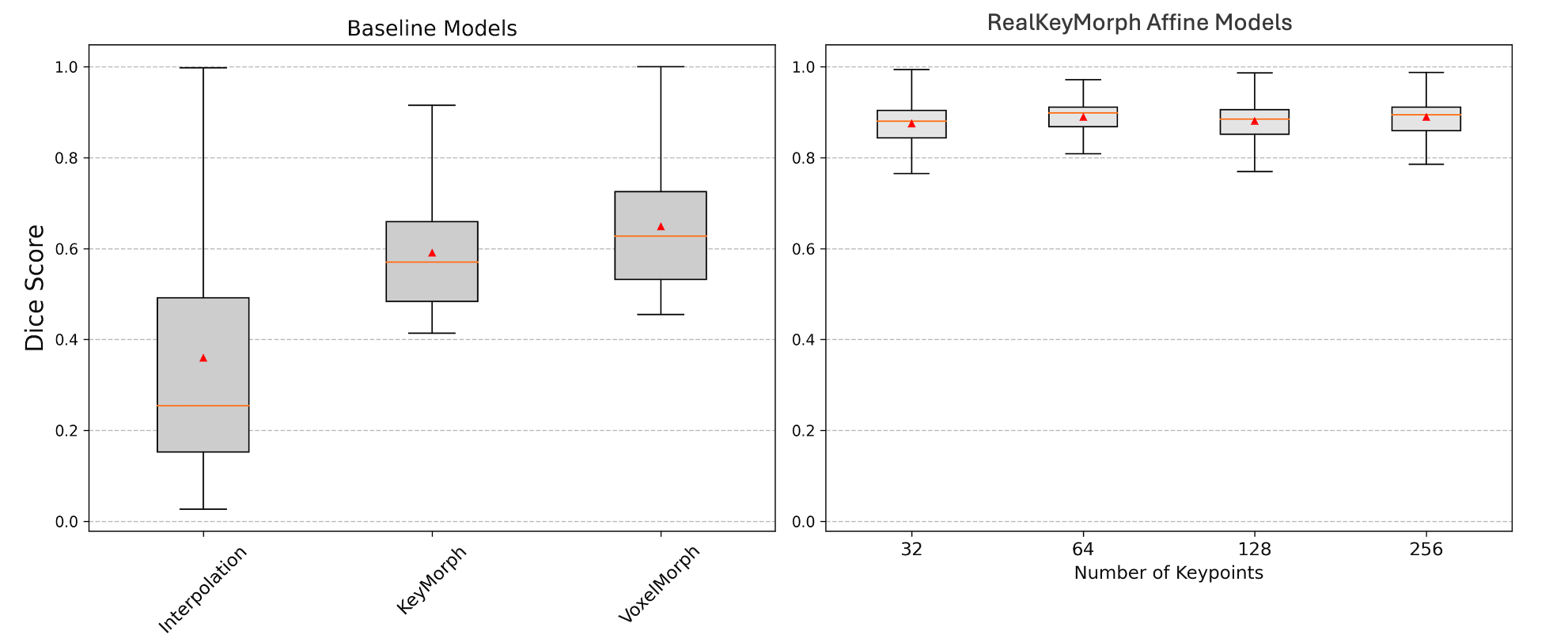}
\caption{Performance of the RealKeyMorph (RKM) model with varying numbers of keypoints on the Brain dataset. The left panel shows the Dice score distribution for baseline models (Interpolation, KeyMorph, VoxelMorph), while the right panel illustrates the performance of the RKM affine models using 32, 64, 128, and 256 keypoints. The box plots display the interquartile range (IQR), with the horizontal orange lines representing the median, and the red triangles indicating the mean Dice score. The best results are achieved with 64 weighted keypoints and $\lambda = 10$, yielding a median Dice score of 89\% and an average Dice score of 89\%.}
\label{fig:Dice Brain}
\end{figure*}

Fig.~\ref{fig:Dice Brain} (right panel) illustrates the model's performance under affine transformation, across varying keypoints configurations. Our findings indicate that using 64 keypoints yields the best performance on the ADNI brain dataset, resulting in a median (IQR) Dice score and HD of 90\% (4\%) and 49 mm (14 mm), respectively. 
The left panel of Fig.~\ref{fig:Dice Brain} also compares the performance of RKM with baseline models (i.e., KeyMorph and VoxelMorph), as well as the initial Dice scores from the interpolated scans before registration. Table~\ref{tab:baselineBrain} show how RKM achieves the best performance on the brain dataset, and how the baseline models compare.

\begin{figure*}[t]
\centering
      \includegraphics[width=0.45\linewidth, height=0.70\linewidth]{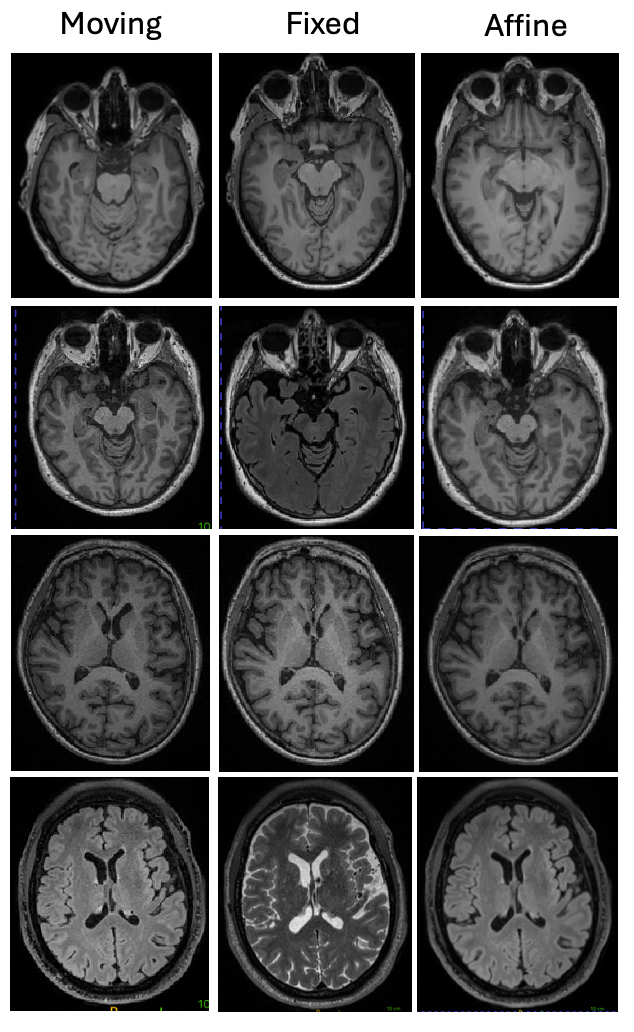}
\caption{Representative subjects from the Brain Dataset, showing fixed and moving images within each subjects, alongside registration results. Each pair includes images with different resolutions. The results demonstrate RKM’s capability for multimodal, resolution-agnostic registration.}
\label{fig:Sample Brain}
\end{figure*}

Fig.~\ref{fig:Sample Brain} shows qualitative results for representative pairs of images with varying MRI sequences within each subject. Each pair includes images with different resolutions and the registration results demonstrate RKM’s capability for multimodal registration.

\begin{table}[htbp] 
    \centering
    \caption{Similarity metrics for the RKM and two baseline models on brain dataset, Median (IQR).}
    \label{tab:baselineBrain}
    \begin{tabular}{lllll}
        \textbf{Model} & \textbf{Keypoints} & \textbf{Dice} & \textbf{HD (mm)} & \textbf{SSIM} \\
        \hline
        VoxelMorph Baseline & NA & 63\% (19\%) & 67 (19) & 55\% (20\%) \\
        KeyMorph Baseline & 512 & 57\% (17\%) & 70 (20) & 43\% (20\%) \\
        RealKeyMorph (Affine) & 64 & 90\% (4\%) & 49 (13) & 79\% (13\%) \\

    \end{tabular}
\end{table}

\section{Discussion}
RKM demonstrates promising results for resolution-agnostic registration of MR stacks and 3D MRI volumes, eliminating interpolation-induced noise and showing potential to enhance MR-based medical diagnostics. This makes RKM a powerful preprocessing tool for image reconstruction tasks involving multiple MR stacks, as it preserves image details during registration and avoids introducing false information into the process. It is also highly valuable for image comparison tasks in applications such as disease analysis and longitudinal monitoring, where subtle anatomical changes can be critical. By eliminating the need for resampling and interpolation, RKM ensures that anatomical information remains intact during the registration step. RKM is also capable of generating a dense set of registrations at test time, guided by a learned set of keypoints. The generalization of these learned keypoints to unseen transformations enables user-controlled registration during inference, allowing adjustment for varying degrees of non-linearity needed for specific applications.


Considering all the capabilities demonstrated by RKM, it should be noted that RKM's performance could be affected by very low signal-to-noise ratios (SNR), significant variations in fields of view, and artifacts caused by the MRI device, patient movement, or breathing. To enhance RKM's robustness to MRI-related noise and artifacts, we plan to incorporate augmentations such as random Gaussian noise and bias fields~\citep{perez2021torchio}, in our future work. Additionally, excessive slice misalignment within 2D stacks and/or slice overlap between sub-stacks in axial views, due to multiple breath-hold acquisitions, can impact RKM's performance. These issues can potentially be mitigated by implementing slice-to-volume registration, providing RKM with cleaner data. By addressing these challenges, we aim to further enhance RKM's performance, making it suitable for a broader range of MRI imaging applications in real clinical settings.





\section{Conclusion}
We presented RealKeyMorph (RKM), a resolution-agnostic deep learning-based registration framework that learns keypoints in real-world coordinates, enabling accurate alignment across MR images with varying resolutions without resampling. By preserving image fidelity and avoiding interpolation artifacts, RKM delivers robust performance on both 2D orthogonal stacks and 3D volumes with varying resolutions. Our experiments demonstrate that RKM outperforms existing methods, offering strong Dice scores and Hausdorff distances across abdominal and brain MRI datasets. Its ability to adapt the degree of non-linearity at inference time adds flexibility for diverse clinical applications, including reconstruction and longitudinal analysis. 


\acks{This research did not receive any specific grant from funding agencies in the public, commercial, or not-for-profit sectors.}

%
\ethics{The work follows appropriate ethical standards in conducting research and writing the manuscript, following all applicable laws and regulations regarding treatment of animals or human subjects.}

\coi{M.R.S. serves as a consultant to Cleerly Health (https://cleerlyhealth.com/). A.Q.W is a postdoctoral researcher at Stanford University.}

\data{The Abdominal dataset underlying this article cannot be shared publicly for the privacy of individuals that participated in the study. The data will be shared on reasonable request to the corresponding author.}

\bibliography{bib-sablab}

\end{document}